%% file: main.tex
\documentclass[11pt,letterpaper]{article}

\input{margins}
\input{preamble}

\usepackage{graphicx}

\usepackage[utf8]{inputenc} 
\usepackage[T1]{fontenc}    
\usepackage{hyperref}       
\usepackage{url}            
\usepackage{booktabs}       
\usepackage{amsfonts}       
\usepackage{nicefrac}       
\usepackage{microtype}      
\usepackage{xcolor}         

\usepackage{amsmath}
\usepackage{amssymb}
\usepackage{url}
\usepackage{booktabs} 
\usepackage{graphicx}
\usepackage{epstopdf}

\usepackage{verbatim}
\usepackage{bm}
\usepackage{array}
\usepackage{multirow}
\usepackage{makecell}
\usepackage{xspace}
\usepackage{pifont}
\usepackage[edges]{forest}
\usepackage{float}
\usepackage{graphicx}
\usepackage{booktabs} 
\usepackage{amssymb} 
\usepackage{pifont} 
\usepackage{tabulary} 
\usepackage{tabularx} 

\graphicspath{{figures/}}
\usepackage{tocloft}
\newcommand{\listdefinitionsname}{List of definitions}
\newlistof{definitions}{dfn}{\listdefinitionsname}

\newtheorem{definition}{Definition}
\usepackage{etoolbox}

\forestset{
  direction switch/.style={
    for tree={edge+=thick, font=\sffamily},
    delay={
      where level>=2{
        grow'=0,
        if content={}{
          coordinate,
          save and restore register={folder indent}{
            folder indent=0,
            folder,
          },
          l sep-/.register=folder indent,
        }{folder},
      }{
        for children=forked edge
      },
    },
  },
}

\usepackage{amssymb,amsmath,amsfonts,bm,epsfig,graphicx,theorem, subfig}
\usepackage{setspace,latexsym,epsf,color,dsfont,epstopdf,wasysym}
\usepackage{cite}
\usepackage{algorithm}
\usepackage{algpseudocode}
\usepackage{mathtools}
\usepackage{graphicx}
\usepackage{multicol}
\usepackage{times,enumitem}
\usepackage{wrapfig,commath}
\usepackage{textcomp}

\let\realtextbf\textbf
\newcommand\answertextbf[1]{%
  \addcontentsline{dfn}{section}{\protect\numberline{\thedefinition}#1}%
  \realtextbf{#1}%
  \let\textbf\real\textbf
}
\usepackage{xparse}
\NewDocumentEnvironment{defn}{o}{
  \let\textbf\answertextbf
  \IfNoValueTF{#1}{\definition}{\definition[#1]}
  }{\enddefinition}

\makeatletter
\renewcommand{\@maketitle}{
\newpage
 \begin{center}%
  {\LARGE \bf\@title \par}%
  \vskip 0.5em%
 {\large \@author \par}%
 \end{center}%
 \par} \makeatother

\newcommand\independent{\protect\mathpalette{\protect\independenT}{\perp}}
\def\independenT#1#2{\mathrel{\rlap{$#1#2$}\mkern2mu{#1#2}}}

\title{\bf \Large A Survey of Out-of-distribution Generalization for Graph Machine Learning from a Causal View}

\date{}
\begin{document}

\author{Jing Ma\\
  {Case Western Reserve University, Cleveland, OH, USA 44106}\\ 
    jing.ma5@case.edu
}
\maketitle

\begin{abstract}
Graph machine learning (GML) has been successfully applied across a wide range of tasks. Nonetheless, GML faces significant challenges in generalizing over out-of-distribution (OOD) data, which raises concerns about its wider applicability. Recent advancements have underscored the crucial role of causality-driven approaches in overcoming these generalization challenges. Distinct from traditional GML methods that primarily rely on statistical dependencies, causality-focused strategies delve into the underlying causal mechanisms of data generation and model prediction, thus significantly improving the generalization of GML across different environments. This paper offers a thorough review of recent progress in causality-involved GML generalization. We elucidate the fundamental concepts of employing causality to enhance graph model generalization and categorize the various approaches, providing detailed descriptions of their methodologies and the connections among them. Furthermore, we explore the incorporation of causality in other related important areas of trustworthy GML, such as explanation, fairness, and robustness. Concluding with a discussion on potential future research directions, this review seeks to articulate the continuing development and future potential of causality in enhancing the trustworthiness of graph machine learning.
\end{abstract}

\input{long}

\bibliographystyle{IEEEtran}
\bibliography{ref}

\section*{Author Biographies}
\textbf{Jing Ma} is a Timothy E. and Allison L. Schroeder Assistant Professor in the Department of Computer \& Data Sciences at Case Western Reserve University (CWRU). Before that, she obtained her Ph.D. in the Department of Computer Science at University of Virginia (UVA) in 2023. She obtained her master's degree and bachelor's degree at Shanghai Jiao Tong University (SJTU). She is broadly interested in machine learning and data mining. Her current research mainly focuses on trustworthy AI (generalization, explanation, fairness, robustness, etc.), causal machine learning, graph mining, AI for social good, and recently large language model. She has won KDD'22 Best Paper Award, and has been selected for the AAAI'24 New Faculty Highlights program.

\end{document}

%% file: margins.tex
\usepackage{calc}
\usepackage[compact]{titlesec}
\titlespacing*{\section}{0pt}{*2}{*2}
\titlespacing*{\subsection}{0pt}{*1.5}{*1.5}
\titlespacing*{\subsubsection}{0pt}{*1}{*1}

\newlength{\myrightmargin}
\newlength{\myleftmargin}
\newlength{\mytopmargin}
\newlength{\mybottommargin}



%% file: preamble.tex
\usepackage{amssymb,amsmath,amsfonts,bm,epsfig,graphicx,theorem}
\newcommand{\comment}[1]{}

\usepackage{comment}

\usepackage{url}

%% file: long.tex

\section{Introduction}
In recent years, graph machine learning (GML), such as graph neural network (GNN) \cite{wu2020comprehensive,zhou2020graph}, has garnered tremendous attention across various research communities, springing up in many high-stakes scenarios such as economic analysis \cite{wang2021review}, scientific discovery \cite{bongini2021molecular}, crime prediction \cite{jin2020addressing}, and pandemic screening \cite{ma2022assessing}. Despite its burgeoning success in various tasks, GML still faces critical challenges, particularly in generalization on out-of-distribution (OOD) data, which casts doubts on its trustworthiness for broader applications. Compared with other data types, generalization on graphs faces unique challenges due to the complex nature of graph structure and the intrinsic dependencies within it, along with the multiple types of distribution shifts in node attributes and graph structure. Therefore, directly employing OOD generalization approaches in other data types often fails on graphs. The unique challenges of GML generalization have spurred a wave of research aimed at enhancing the generalization capabilities of GML through diverse approaches \cite{li2022out}.

Among existing studies in GML generalization, causality-involved GML methods \cite{jiang2023survey} have made eye-catching progress. Unlike conventional GML approaches that primarily leverage statistical dependencies for downstream tasks, recent studies \cite{ma2022learning,kaddour2022causal} suggest that causality often plays a more pivotal role in understanding the underlying data generation mechanisms. For instance, traditional GML techniques tend to overfit and rely excessively on spurious correlations \cite{arjovsky2019invariant},  which, while effective in some scenarios, often fail to capture the causal mechanisms that fundamentally govern the relationships within the data,
thereby undermining the models' ability to generalize across new data domains. Causality \cite{pearl2009causality}, with its focus on causal relationships rather than merely statistical dependencies, thereby can naturally discern true causal mechanisms and eliminate spurious correlations.
This motivation has led to a growing enthusiasm for integrating causality into GML frameworks.

In this paper, we systematically review recent advancements in causality-involved GML generalization, covering their objectives, technologies, and effectiveness from different angles. More specifically, we first introduce the principal concepts and intuition that employ causality to enhance the generalization of graph models. We then categorize existing methods into distinct branches, detailing their techniques and exploring their interconnections. Additionally, we broaden our discussion to include the application of causality in other critical aspects of trustworthy graph machine learning. Finally, we highlight unresolved issues, articulate ongoing challenges, and outline prospective directions for future research in this rapidly evolving field.

\textbf{Differences from existing surveys.} There have been a couple of surveys in related areas, including graph OOD generalization \cite{li2022out}, causality-inspired graph neural networks \cite{jiang2023survey,job2023exploring}, causality learning in graphs \cite{ma2022learning}, causal machine learning \cite{kaddour2022causal}, and trustworthy GML \cite{wu2022survey,yuan2022explainability,dong2023fairness}. The surveys closest to us are \cite{li2022out} and \cite{jiang2023survey}, but our work distinguishes itself from existing ones in the following aspects: (1) \textbf{Main Focus.} Our survey mainly centers on OOD generalization within GML from a causal perspective, a topic that other surveys have not primarily addressed. (2) \textbf{Organization.} We have structured the review of existing works differently, offering a unique layout that enhances understanding and integration of the mentioned materials. (3) \textbf{Timeline.} We provide coverage of the most recent advancements and discussions in this field. To be best of our knowledge, there has not been a comprehensive survey focusing on this unique topic.

The general organization of this survey is presented below:
\begin{itemize}
    \item \textbf{Overview and preliminaries.} In Section \ref{sec:preliminaries}, we introduce the background knowledge and give an overview of this area. First, we will present the key concepts in causality learning, the common scenarios and tasks of GML, and the generalization issues. We also introduce the motivations and challenges of incorporating causality into graph machine learning.
    \item \textbf{Method review.} In Section \ref{sec:method}, we categorize existing approaches of causality-involved GML generalization into different groups and introduce their techniques and connections.
    We will start with the basic intuition of these studies, and then review the related frameworks and technologies. These studies mainly cover invariant learning, causal model-based methods, and stable learning.
    \item \textbf{Connection to other trustworthy domains.} In Section \ref{sec:other domains}, we extend our discussion to the use of causality in other related areas of trustworthy GML, such as explanation, fairness, and robustness. We explore the intrinsic connections between these domains to offer readers a more comprehensive and expandable sight for the whole picture of the related research areas.
    \item \textbf{Future work.} In Section \ref{sec:future}, we summarize current efforts in causality-involved GML generalization. Furthermore, we look forward and outline promising future research directions.
\end{itemize}

\section{Preliminaries} 
\label{sec:preliminaries}
\subsection{Causality and Causal Inference}
In this section, we provide important background knowledge on causal inference. We start with the key tasks and concepts in causal inference, introducing the notations, definitions, and frameworks. 
Generally, causal inference aims to investigate the causality between different data variables. 

\begin{defn}
\textbf{(Structural causal model (SCM)).} Structural causal model \cite{pearl2009causality} is a widely adopted framework to model causal relationships. An SCM can be defined with a triplet of sets $(\mathcal{U}, \mathcal{V}, \mathcal{F})$, here, $\mathcal{U}$ is a set of exogenous variables, $\mathcal{V}$ is a set of endogenous variables, and $\mathcal{F}=\{f_1(\cdot),f_2(\cdot),...,f_{|\mathcal{V}|}(\cdot)\}$ is a set of functions (known as \textit{structural equations}) that describe the causal relationships between variables. For each $V \in \mathcal{V}$, there is a structural equation $V=f_V(pa_V, U_V)$, where $pa_V\subseteq \mathcal{V}\setminus V$, $U_V\subseteq U$ are variables that directly cause $V$.
\end{defn}

Each SCM is associated with a causal graph, usually, it is a directed acyclic graph (DAG) with variables represented as nodes and causal relationships represented as directed edges. The conditional independence relationships between variables are straightforwardly reflected by the causal graph. Different from dependencies, causal relations only spread through directed paths whose edges are in the same direction. 
On causal graphs, there are three types of basic junctions: \textit{chain} (e.g., $X\rightarrow Z \rightarrow Y$, here $X$ has causal effects on $Y$ through a \textit{mediator} $Z$), \textit{fork} (e.g., $X\leftarrow Z \rightarrow Y$, here $Z$ serves as a \textit{confounder} which brings non-causal dependency between $X$ and $Y$), and \textit{collider} (e.g., $X\rightarrow Z \leftarrow Y$, here $Z$ is a collider bringing non-causal dependency between $X$ and $Y$ when conditioning on $Z$).
One of the key challenges in causal inference is to identify the causal relationships or effects out of all the statistical dependencies. The gold standard approach of causal inference --- randomized controlled trials (RCTs) are often infeasible or unethical to practice in the real world. Here, we introduce a couple of other commonly used approaches and their related concepts. A foundational concept in SCM is the do-operator $do(\cdot)$, which stands for an intervention. Based on this, we have the following definition of causal effect in an average case:

\begin{defn} \textbf{(Average treatment effect).} The average causal effect of a certain \textit{treatment} (a.k.a. cause) $T$ (for simplicity, we usually assume it is a binary variable) on an outcome $Y$ can be formalized as follows:
$$ATE = \mathbb{E}[Y|do(T=1)] - \mathbb{E}[Y|do(T=0)]$$
\end{defn}

For a pair of treatment $T$ and outcome $Y$, there often exist backdoor paths that possibly bring non-causal dependencies for them. A \textit{backdoor path} neither is a directed path nor contains any collider. Typical ways for unbiased causal effect estimation include \textit{backdoor adjustment}, \textit{frontdoor adjustment}, and \textit{instrumental variable (IV)}-based approaches.

\begin{defn}\textbf{(Backdoor adjustment).} Under certain basic causal assumptions \footnote{These assumptions include the Modularity assumption and Positivity assumption.}, for a pair of treatment $T$ and outcome $Y$, if there are variables $Z$ that (1) block all the backdoor paths between $T$ and $Y$, (2) do not contain any descendants of $T$, we have
$$
P(Y|do(t))=\sum\nolimits_Z P(Z)P(Y|t,Z)
$$
\end{defn}

In certain cases (e.g., when hidden confounders exist), it is difficult to find an observed variable set $Z$ for backdoor adjustment. Alternatively, we can search for other variable sets which either meet the frontdoor criterion, or serve as instrumental variables. 

\begin{defn} \textbf{(Frontdoor criteron).} A set of variables $M$ satisfies the frontdoor criterion relative to $T$ and $Y$ if: (1) $M$ completely mediates the effect of $T$ on $Y$; (2) There is no unblocked backdoor path from $T$ to $M$; (3) All backdoor paths from $M$ to $Y$ are blocked by $T$.
\end{defn}

\begin{defn} \textbf{(Instrumental variable (IV)).} Given treatment $T$ and outcome $Y$, a variable $I$ can serve as an instrumental variable if it satisfies the following conditions: (1) (Relevance) $I$ is relevant to the treatment $T$; (2) (Exclusion restriction) The causal effect from $I$ to $Y$ is mediated by $T$; (3) (Instrumental unconfoundedness) There is no unblocked backdoor path from $I$ to $Y$.
\end{defn}

For variables satisfying frontdoor criterion, we can use frontdoor adjustment \cite{pearl2009causality} to estimate causal effect. Similarly, for instrumental variables (IVs), causal effects can be determined using appropriate IV-related methodologies \cite{baiocchi2014instrumental}.


\tikzstyle{my-box}=[
    rectangle,
    draw=black,
    rounded corners,
    text opacity=1,
    minimum height=1.5em,
    minimum width=5em,
    inner sep=2pt,
    align=center,
    fill opacity=.5,
]
\tikzstyle{leaf}=[
    my-box, 
    minimum height=1.5em,
    text=black,
    align=left,
    font=\normalsize,
    inner xsep=4pt,
    inner ysep=4pt,
]
\tikzstyle{leaf_1}=[
    my-box, 
    minimum height=1.5em,
    text=black,
    align=left,
    font=\normalsize,
    inner xsep=4pt,
    inner ysep=4pt,
]
\begin{figure*}[t!]
    \centering
    \resizebox{\textwidth}{!}{
        \begin{forest}
            forked edges,
            for tree={
                grow=east,
                reversed=true,
                anchor=base west,
                parent anchor=east,
                child anchor=west,
                base=left,
                font=\large,
                rectangle,
                draw=black,
                rounded corners,
                align=left,
                minimum width=4em,
                edge+={darkgray, line width=1pt},
                s sep=3pt,
                inner xsep=4pt,
                inner ysep=3pt,
                line width=0.8pt,
                ver/.style={rotate=90, child anchor=north, parent anchor=south, anchor=center},
            },
            where level=1{
                text width=13.0em,
                font=\normalsize,
                inner xsep=4pt,
                }{},
            where level=2{
                text width=11.0em,
                font=\normalsize,
                inner xsep=5pt,
            }{},
            where level=3{
                text width=13em,
                font=\normalsize,
                inner xsep=4pt,
            }{},
            where level=4{
                text width=12em,
                font=\normalsize,
                inner xsep=4pt,
            }{},
            [
                Causality-involved GML OOD generalization, ver
                [   
                    Invariant learning (Section \ref{sec:invariant}) 
                    [
                        Node-level invariance
                        [
                            EERM \cite{EERM}{,} INL \cite{INL}{,} FLOOD \cite{FLOOD}{,} \\
                            SILD \cite{SILD}{,} DIDA \cite{DIDA}
                            , leaf, text width=15em
                        ]
                    ]
                    [
                        Graph-level invariance
                        [
                            GIL \cite{GIL}{,} DIR \cite{DIR}{,} GSAT \cite{GSAT}{,} \\
                            RGCL \cite{RGCL}{,} CIGA \cite{CIGA}{,} GALA \cite{GALA}{,}\\
                            DisC \cite{DisC_2022}{,} IGM \cite{IGM}{,} LECI \cite{LECI_2024}{,} \\
                            MoleOOD \cite{MoleOOD}{,} iMoLD \cite{iMoLD}  
                            , leaf, text width=15em
                        ]
                    ]
                ] 
                [
                    Causal modeling (Section \ref{sec:causalmodel})
                    [
                        Backdoor adjustment
                        [
                            CAL \cite{CAL_2022}{,} CAL+ \cite{CAL+_2024}{,} CaNet \cite{CaNet_2024}
                            , leaf, text width=15em
                        ]
                    ]
                    [
                        Frontdoor adjustment
                        [
                            DSE \cite{DSE_2022}
                            , leaf, text width=15em
                        ]
                    ]
                    [
                        Instrumental variable
                        [
                            RCGRL \cite{RCGRL_2023}
                            , leaf, text width=15em
                        ]
                    ]
                    [
                        Graph models inspired \\causal models
                        [
                            E-invariant GR \cite{E_invariant_2021}{,} gMPNN \cite{gMPNN_2022}
                            , leaf, text width=15em
                        ]
                    ]
                    [
                        Counterfactual reasoning
                        [
                            CFLP \cite{CFLP_2022}
                            , leaf, text width=15em
                        ]
                    ]
                ] 
                [
                    Stable learning (Section \ref{sec:stable})
                    [
                        OOD-GNN \cite{OOD-GNN_2022}{,} StableGNN \cite{StableGNN_2023}{,} DGNN \cite{DGNN_2022}{,} L2R-GNN \cite{L2R-GNN_2023}
                        , leaf, text width=28em
                    ]
                ]
            ]
        \end{forest}
    }
    \caption{The representative methods for causality-involved GML OOD generalization.}
    \label{fig:methods}
\end{figure*}
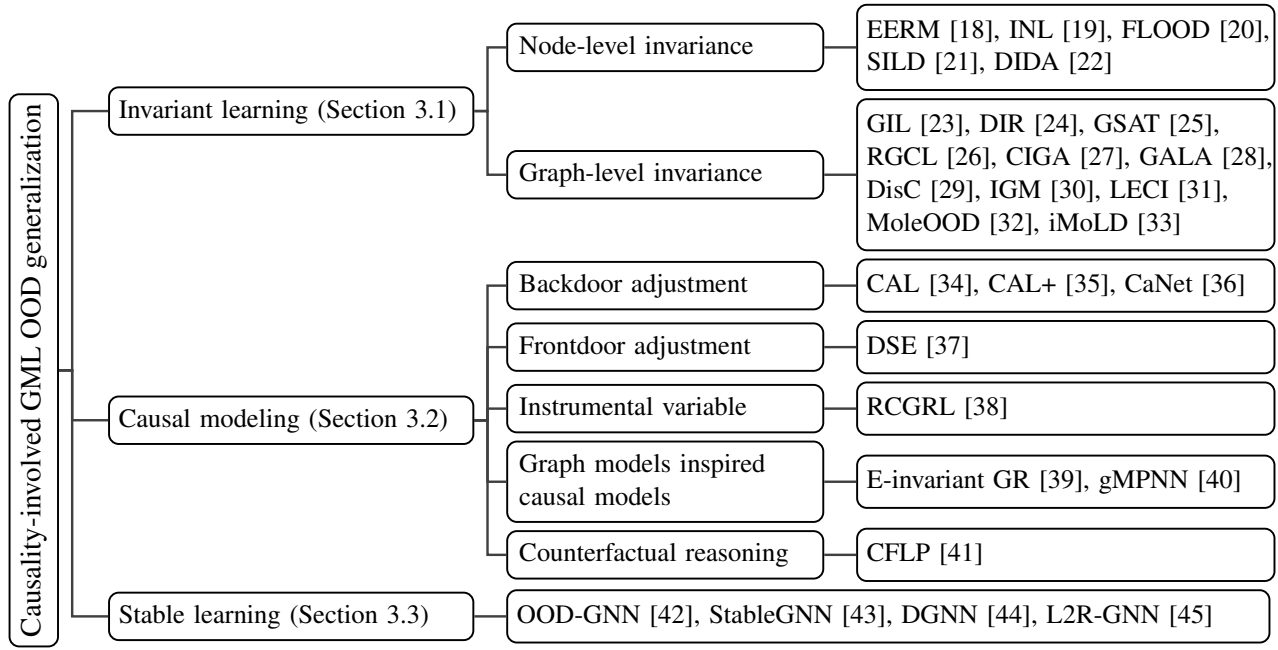

\subsection{Causality and Graph Machine Learning}
A graph can be denoted by $G=(X,A)$, where $X\in \mathcal{R}^{n\times d}$ denotes the node features and $A\in \{0,1\}^{n\times n}$ is an adjacency matrix representing the graph edges. Here, $n$ is the number of nodes, and $d$ is the feature dimension. GML involves a variety of tasks related to graphs with the prediction target $Y$ in different granularities, such as local-level prediction tasks like node classification and link prediction, as well as graph-level tasks like graph classification. 
In GML, models are specifically designed to handle the unique challenges posed by graph data, e.g., capturing the dependencies and relationships between nodes. Prominent models in this domain include many branches extending traditional neural network architectures to operate on graph structures, such as the representative graph convolutional network (GCN) \cite{kipf2016gcn}, graph attention network (GAT) \cite{velivckovic2017graph}, graph variational autoencoder \cite{kipf2016variational}, and graph transformer \cite{yun2019graph}. These models have achieved remarkable performance in graph-related tasks, but there are still concerns regarding their trustworthiness in different aspects. One of the key concerns is the generalization ability of these GML methods, as many of them easily fall short under distribution shift (i.e., $P^{train}(G,Y)\ne P^{test}(G,Y)$) in OOD data. Noticeably, such distribution shift may come from both node features and the complicated graph structure.

In recent years, many researchers have realized the necessity of incorporating causality into machine learning to enhance model trustworthiness. As aforementioned, causality distinguishes itself from other generalization methods by its inherent ability to uncover causal relationships within data, which is particularly valuable in GML. However, implementing causality in GML presents multifaceted challenges. Firstly, causality within graphs is intrinsic; generalizing on graphs is inherently complex and often involves obscure causal relationships across domains. Secondly, causal assumptions that hold in traditional settings may not directly apply to graphs, complicating the adaptation of standard causal methods to this context. Addressing these challenges, there has been a growing interest in exploring the interconnection between causality and GML, yielding insightful contributions and promising prospects for the future.









\begin{table*}[]
\begin{tabular}{llcccccc}
\toprule
\textbf{Method}      & \textbf{Shift} & \textbf{Task}      & \textbf{E known}   & \textbf{Single E} & \textbf{SCM} & \textbf{Theory}                         & \textbf{Application}  \\
\midrule
\textbf{EERM} \cite{EERM}        & $A,X$                                 & Node         & No    & Yes & No           & Yes                & General                  \\
\textbf{INL}   \cite{INL}      & $A,X$                                 & Node         & No & No        & No           & Yes                & General                  \\
\textbf{FLOOD}  \cite{FLOOD}     & $A,X$                                 & Node         & No    & No               & No           & No                & General                  \\
\textbf{GIL}  \cite{GIL}       & $A,X$                                 & Graph         & No     & No               & No           & Yes                & General                  \\
\textbf{DIR}  \cite{DIR}       & $A,X$                                 & Graph         & No         & No               & Yes           & Yes                & General                  \\
\textbf{GSAT}  \cite{GSAT}      & $A,X$                                 & Graph        & No         & No               & Yes           & Yes                & General                  \\
\textbf{RGCL}  \cite{RGCL}      & $A,X$                                 & Graph         & No         & No               & No           & No                & General                  \\
\textbf{CIGA}   \cite{CIGA}     & Size, $A,X$                             & Graph         & No         & No               & Yes           & Yes                & General                  \\
\textbf{GALA}   \cite{GALA}     & Size, $A,X$                           & Graph         & No         & No               & Yes           & Yes                & General                  \\
\textbf{DisC}   \cite{DisC_2022}     & $A,X$                                 & Graph         & No         & No               & Yes           & No                & General            \\
\textbf{IGM}  \cite{IGM}       & Degree, size, $A,X$                      & Graph         & No         & No               & No           & No                & General                  \\
\textbf{LECI}   \cite{LECI_2024}     & $A,X$                                 & Graph         & Yes         & No        & Yes           & Yes                & General            \\
\textbf{SILD}  \cite{SILD}      & (Dynamic) $A,X$             & Node/link       & No         & No        & No           & Yes                & Dynamic graph     \\
\textbf{DIDA}   \cite{DIDA}     & (Dynamic) $A,X$                         & Graph & No         & No               & No           & No                & Spatio-temporal    \\
\textbf{MoleOOD}  \cite{MoleOOD}   & Size, $A$                            & Graph         & No         & No               & No           & Yes                & Molecular          \\
\textbf{iMOLD}   \cite{iMoLD}    & Size, $A$                            & Graph         & No         & No               & No           & No                & Molecular          \\
\midrule
\textbf{CAL} \cite{CAL_2022}   & $A,X$                                 & Graph         & No         & No               & Yes           & Yes                & General                  \\
\textbf{CAL+}  \cite{CAL+_2024}  & $A,X$                                 & Graph         & No        & No               & Yes           & Yes                & General                  \\
\textbf{CaNet}  \cite{CaNet_2024}     & $A,X$                                 & Node      & No         & No               & Yes           & No                & General                  \\
\textbf{DSE}    \cite{DSE_2022}     & $A$                                  & Graph         & No         & No               & Yes           & Yes                & General                  \\
\textbf{RCGRL}   \cite{RCGRL_2023}    & $A$                                  & Graph         & No         &    No             & No           & Yes                & General                  \\
\textbf{E-invariant} \cite{E_invariant_2021}&  Size, $X$ & Graph         & No         &  No               & Yes           & Yes                & General            \\
\textbf{gMPNN} \cite{gMPNN_2022}      & Size, $X$                           & Link      & No         & No               & Yes           & Yes                & General            \\
\textbf{CFLP}  \cite{CFLP_2022}      & $A$                                  & Link      & No         & No               & Yes           & No                & General \\
\midrule
\textbf{OOD-GNN }\cite{OOD-GNN_2022}    & Size, $A,X$                             & Graph         & No         &     No            & No           & Yes                & General            \\
\textbf{StableGNN} \cite{StableGNN_2023}   & $A,X$                                 & Graph         & No         & No               & Yes           & No                & General            \\
\textbf{DGNN} \cite{DGNN_2022}        & $A,X$                                 & Node      & No         &     No            & Yes           & Yes                & General                  \\
\textbf{L2R-GNN} \cite{L2R-GNN_2023}    & $A$                                  & Graph         & No         & No               & No           & No                & General  \\  
\bottomrule
\end{tabular}
\caption{Comparison of existing causality-involved GML generalization methods in different aspects, including applicable shift type (including $A$=graph structure, $X$=node features, graph size, and node degree), prediction task, whether environment labels are known, whether the method assumes a single training environment, whether the method is supported by an SCM, whether there is a theoretical guarantee for the method, and application scenario.}
\label{tab:method}
\end{table*}

\section{Methodologies}
\label{sec:method}
In this section, we will introduce the principles and methodologies of leveraging causality for GML generalization. We categorize the mainstream of works into invariant learning, causal modeling, and stable learning. Fig. \ref{fig:methods} shows an overview of the categorization for these methods. A more detailed comparison of these methods is in Table \ref{tab:method}.

\subsection{Invariant Learning}
\label{sec:invariant}
Invariant learning targets on capturing the relations that are invariant across different domains. It is motivated by the fact that spurious correlations often vary under distribution shifts. The general principle of invariant learning is improving generalization by extracting the invariant factors to make predictions, while the spurious correlations are filtered out. The formal assumptions of invariant learning slightly vary in different works, but generally, there should exist invariant factors $\Phi(G)$ from input $G$, that $P^e(Y|\Phi(G))$ remains stable in different domain $e$. Although invariant learning is not explicitly driven by causal inference, much literature has discussed their intrinsic connections \cite{wang2022unified,mitrovic2020representation}. In general, in a causal view, the direct causes for the label $Y$ should have invariant relationships in different domains. The idea of invariant learning is first proposed for tabular data, including representative methods such as invariant risk minimization (\textbf{IRM}) \cite{arjovsky2019invariant} and \textbf{EIIL} \cite{creager2021environment}. However, on graphs, directly applying these methods often results in unsatisfying performance. Invariant learning on graphs often extracts a subgraph as a rationale that generalizes across domains. Accordingly, many graph-specific invariant learning methods have been proposed in recent years, and take one of the mainstreams in OOD generalization for GML.

\textbf{Node-level invariance learning.} Explore-to-Extrapolate Risk Minimization (\textbf{EERM}) \cite{EERM} uses multiple adversarial context generators to simulate (virtual) environments even under a single (real) environment, and a GNN model is trained by minimizing the mean and variance of risks from these simulated environments. Different from the single environment setting in EERM, Li et al. \cite{INL} argue that nodes are often from multiple latent environments in the real world, and propose an approach \textbf{INL} that can infer node environments with a contrastive modularity-based graph clustering method and learn invariant node representations. 
A recent unique work Flexible invariant Learning framework for Out-Of-Distribution generalization on graphs (\textbf{FLOOD}) \cite{FLOOD} combines invariant learning and bootstrapped learning. It first constructs multiple training environments based on data augmentation, then adopts a bootstrapped learning module. In this way, their encoder is more flexible than traditional invariant encoders as it can be refined on the test set for better generalization. 

\begin{figure*}[t!]
\centering
\includegraphics[width=0.98\textwidth,height=3.4in]{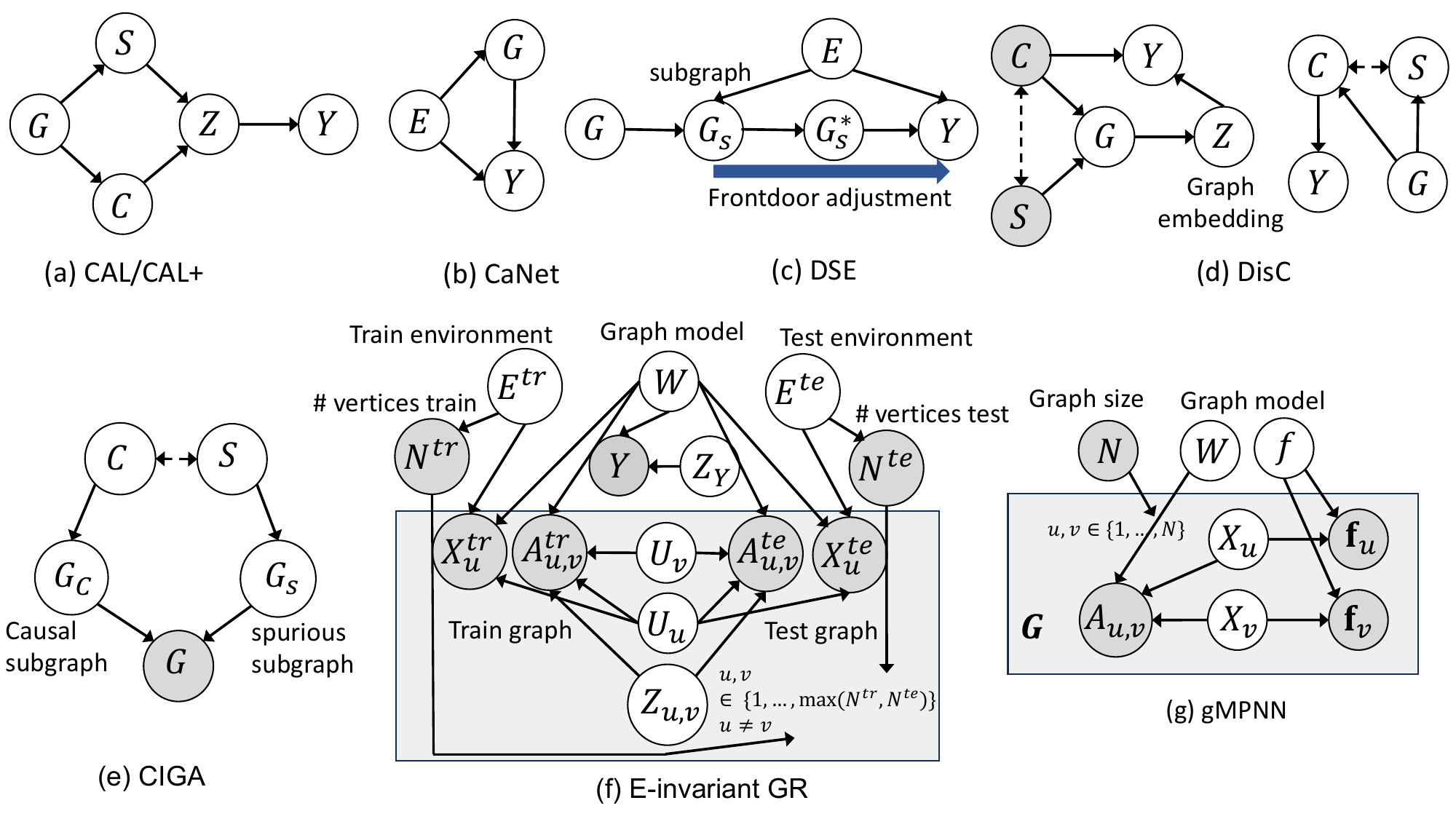}
\caption{An overview of causal graphs employed in representative GML OOD generalization methods, including CAL \cite{CAL_2022}, CAL+ \cite{CAL+_2024}, CaNet \cite{CaNet_2024}, DSE \cite{DSE_2022}, DisC \cite{DisC_2022}, CIGA \cite{CIGA}, E-invariant GR \cite{E_invariant_2021}, and gMPNN \cite{gMPNN_2022}. $G$, $C$, $S$, $Y$, $Z$, and $E$ denote graphs, causal variables, spurious variables, prediction labels, graph representations, and environments, respectively. $G_C$ and $G_S$ in (e) represent the causal subgraph and spurious subgraph; while $G_S$ and $G_S^*$ in (c) represent an explanatory subgraph and a surrogate subgraph, respectively. For those causal graphs with nodes in different colors, observed/unobserved variables are in grey/white. Dashed lines show unknown correlations.}
\label{fig:causal_model}
\end{figure*}

\textbf{Graph-level invariance learning.} Graph Invariant Learning (\textbf{GIL}) \cite{GIL} is a GNN-based model that identifies the invariant subgraph for graph classification tasks. It is the first work of invariant graph representation learning under mixed latent environments without the supervision of environment labels. Discovering Invariant Rationales (\textbf{DIR}) \cite{DIR} is an algorithm that infers invariant causal parts by conducting causal interventions, but it needs a complicated iterative process to break and assemble subgraphs during training. Another more straightforward method Graph Stochastic Attention (\textbf{GSAT}) \cite{GSAT} is based on the information bottleneck principle. It learns invariant subgraphs by learning stochasticity-reduced attention. 
Li et al. \cite{RGCL} propose Rationale-aware Graph Contrastive Learning (\textbf{RGCL}), which combines invariant rationale discovery with graph contrastive learning to improve generalization and interoperability. \textbf{CIGA} \cite{CIGA}, with a supported causal graph shown in Fig.~\ref{fig:causal_model}(e), proposes an information-theoretic objective to extract invariant subgraphs with a theoretical guarantee to handle distribution shift under different SCMs. In a follow-up work, Chen et al. \cite{GALA} analyze the failure cases of existing methods such as CIGA, and introduce minimal assumptions for feasible invariant graph learning. They propose a framework \textbf{GALA} with provable invariant subgraph identifiability for OOD generalization. \textbf{DisC} \cite{DisC_2022}, inspired by a causal graph shown in Fig.~\ref{fig:causal_model}(d), disentangles the given graph into a causal substructure and a bias substructure. The disentanglement is conducted with a parameterized edge mask generator. Then two GNNs are trained to encode the causal and bias substructures respectively into their representations trained with causal/bias-aware loss functions. To further decorrelate causal and bias variables, DisC also generates unbiased counterfactual training samples. For all of the invariant learning biased methods, a common conclusion is that the effectiveness of invariant learning is greatly dependent on the variety of environments. Realizing this, a co-mixup strategy \textbf{IGM} \cite{IGM} is proposed, which jointly adopts environment mixup and invariant mixup to generate sufficiently diverse environments. Many existing graph OOD generalization methods either do not assume the existence of environment labels, or do not fully exploit them. Differently, \textbf{LECI} \cite{LECI_2024} is proposed to utilize pre-collected environment information for graph-specific OOD generalization. It discovers causal invariant subgraphs by leveraging the two causal independence properties regarding label and environment: $E\independent G_C$ and $Y\independent G_S$ and designs an adversarial learning strategy to jointly optimize these two causal independence properties. Here, $E$ is the environment, $G_C$ and $G_S$ are the causal subgraph and spurious subgraph, respectively.

\textbf{Domain-specific invariant learning.} \textbf{SILD} \cite{SILD} is the first work to study distribution shifts on dynamic graphs in the spectral domain, it captures invariant and variant spectral patterns with disentangled spectrum masks for both node classification and link prediction tasks. 
Dynamic graph Attention network (\textbf{DIDA}) \cite{DIDA} handles complex spatio-temporal distribution shifts in dynamic graphs by using a spatio-temporal attention network to identify variant and invariant spatio-temporal patterns. In molecular representation learning (MRL), MoleOOD \cite{MoleOOD} is the first work that formulates the OOD problem in MRL that leverages the invariance principle. It includes an environment inference model and guides their encoder in learning environment-invariant molecular substructures.
In a similar context, invariant Molecular representation in Latent Discrete space (\textbf{iMoLD}) \cite{iMoLD} separates the molecular graph representation into invariant and spurious parts with a GNN scorer, and uses a task-agnostic self-supervised learning objective to further improve invariance identification. This approach does not require environment inference in other works like MoleOOD and GIL, and thus avoids the additional assumptions and knowledge on environment distrbutions required in environment inference.

\subsection{Causal Modeling}
\label{sec:causalmodel}
Inspired by studies exploring causal relationships within graphs \cite{ma2022learning,jiang2023survey}, a notable line of research has emerged that explicitly constructs SCMs on graphs to enhance GML OOD generalization. This research typically relies on predefined assumptions about the underlying SCM, which are illustrated through clear causal graphs that illustrate the causal relationships among the graph, labels, causal features, and spurious features. An overview of these commonly used causal graphs is presented in Fig.~\ref{fig:causal_model}. These methods often employ traditional causal inference principles to reduce spurious correlations and improve generalization. Based on the key techniques, we categorize these methods into several branches: backdoor adjustment, frontdoor adjustment, instrumental variable-based methods, and those related specifically to graph modeling. 

\textbf{Backdoor adjustment.} Causal Attention Learning (\textbf{CAL}) \cite{CAL_2022} is based on the causal assumption that there exist shortcut features that serve as confounders between the causal features and graph prediction. Under this assumption, CAL employs attention modules to estimate the causal features and shortcut features of the input graph, and conduct backdoor adjustment to mitigate the spurious correlations led by the backdoor path $C \leftarrow G \rightarrow S \rightarrow Z \rightarrow Y$, shown in Fig. \ref{fig:causal_model}(a). A follow-up work \textbf{CAL+} \cite{CAL+_2024} generally inherits the idea of CAL, but it further enhances the method with a memory bank to improve the diversity of shortcut feature samplings and a prototype module to enhance the consistency of intra-class causal features. For node-level distribution shift, \textbf{CaNet} \cite{CaNet_2024} uses a straightforward causal graph shown in Fig. \ref{fig:causal_model}(b), relying on backdoor adjustment and variational inference. It counteracts the confounding bias by collaboratively training an environment estimator and a GNN predictor. 

\textbf{Frontdoor adjustment and instrumental variable.} Wu et al. \cite{DSE_2022} also attribute the OOD distribution shift to the confounder effect. As shown in Fig. \ref{fig:causal_model}(c), they propose a method Deconfounded Subgraph Evaluation (\textbf{DSE}) which introduces a surrogate $G_S^*$ between the explanatory subgraph $G_S$ and model prediction $Y$ to mitigate confounding bias via frontdoor adjustment. The surrogates are generated by a generative model based on a conditional variational graph auto-encoder. Another approach \textbf{RCGRL} \cite{RCGRL_2023} eliminates confounding bias by generating instrumental variables (IV) under unconditional moment restrictions. On graphs, the conditions of instrumental variables are often hard to satisfy, instead, RCGRL proposes an active IV generation approach that transfers the conditional moment restrictions into unconditional ones with theoretical support. 

\textbf{Graph models inspired causal models.} \textbf{E-invariant GR} \cite{E_invariant_2021} constructs a fine-granularity causal graph (shown in Fig. \ref{fig:causal_model}(f)) including graphon, label, training/text environment, node attributes and edges of training/test graph, and number of nodes. With this causal graph, it aims to learn environment-invariant graph representations that are generalizable to shifts in graph size and node attributes. The invariant representations are learned based on the stability of subgraph densities in graphon random graph models \cite{lovasz2006limits}. This unique work incorporates Stochastic Block Models
(SBMs) \cite{diaconis1981statistics,snijders1997estimation} and graphon random graph models \cite{airoldi2013stochastic,lovasz2006limits} inside its causal model. A study in a similar setting \cite{gMPNN_2022} extends to link prediction problem by proposing a new family of structural pairwise embeddings \textbf{gMPNN}, with its causal graph shown in Fig. \ref{fig:causal_model}(g).

\textbf{Counterfactual reasoning.} In the context of causal inference, the term "\textit{counterfactual}" stands for a hypothetical scenario that deviates from actual events. For instance, one might ask, "Would the label have changed if a specific subgraph had been altered?". In this line of research, \textbf{CFLP} \cite{CFLP_2022} employs counterfactual reasoning for OOD link prediction. It focuses on the causal relationships between the graph structure and the existence of links, i.e., “would the link still exist if the graph structure became different?”. This method is achieved by training GNN-based link predictors to predict both actual (factual) links and counterfactual links.



\subsection{Stable Learning} 
\label{sec:stable}
Stable learning \cite{StableGNN_2023,cui2022stable}, with its primary goal of learning a model that can perform uniformly well in any environment, originates from the sampling reweighting or covariate balancing strategies in causal effect estimation \cite{imai2014covariate}. More specifically, many traditional causal inference methods estimate causal effects by using sampling reweighting to assign sample weights that balance the distribution of covariates across different treatment groups. In stable learning, each input feature is treated as a treatment, while other variables are considered covariates. It focuses on learning weights to decorelate the features, thereby balancing the covariate distribution with respect to each feature treated as a treatment. In this way, the correlation between each feature (treatment) and the prediction label (outcome) results from a direct causal effect. Then a predictive model based on correlation can achieve better generalization across varied data environments.

In this area of studies, \textbf{OOD-GNN} \cite{OOD-GNN_2022} adopts a nonlinear graph representation decorrelation method that leverages random Fourier features. It reduces the statistical dependencies between relevant and irrelevant graph representations by iteratively optimizing the sample weights and the graph encoder. This approach promotes model generalization against multiple shift types, including graph size, node attribute, and graph structure. Different from OOD-GNN which learns a single embedding for each graph, \textbf{StableGNN} \cite{StableGNN_2023} extracts high-level representations for subgraphs with graph pooling layers from the input graph. Based on these high-level representations, StableGNN designs a causal variable distinguishing regularizer to de-bias the training distribution through sample weight learning. Similarly, Debiased GNN (\textbf{DGNN}) \cite{DGNN_2022} devises a differentiated decorrelation regularizer for debiasing at the node level. Later on, researchers argue that methods like StableGNN and OOD-GNN may suffer from the overly aggressive objective that eliminates the dependencies between all the variables across the graph representations, leading to an excessively small sample size. With this motivation, they propose \textbf{L2R-GNN} \cite{L2R-GNN_2023} which first clusters the variables in graph representations based on the correlation stability, and then only learns weights to eliminate correlations between variables across different clusters, instead of removing correlations between any pair of variables.

\subsection{Discussion}
Despite the categorizations outlined above, many methods demonstrate deep interconnections and underlying equivalencies. For example, several invariant learning techniques, such as CIGA \cite{CIGA} and Disc \cite{DisC_2022}, are explicitly underpinned by structural causal models, as shown in Fig.~\ref{fig:causal_model}, positioning them at the intersection of multiple categories. Additionally, some other techniques are also closely related to the approaches discussed, including disentangled representation learning \cite{DisC_2022}, representation decorrelation \cite{OOD-GNN_2022,StableGNN_2023}, causal interventions \cite{CaNet_2024}, counterfactual reasoning \cite{CFLP_2022}, and data augmentation \cite{DisC_2022,GREA_2022}. These connections highlight the complex, often overlapping landscape of methodologies within this field.

\section{Connection to Other Domains in Trustworthy Graph Machine Learning}
\label{sec:other domains}
Even though this survey mainly focuses on causality-based generalization for GML, it is worth mentioning that many principles and technologies in this area closely connect to other domains of trustworthy GML. Here, we give a conceptual overview of these intrinsic connections.

\subsection{Explanation}
Although explanation has many different definitions, in graphs, there are generally two main categories for GML explanation, including those that aim to identify the rationale that contributes most to prediction (\textbf{factual explanation}) with the question "\textit{what contributes most to the prediction}?", and counterfactuals that can achieve a certain desired outcome (\textbf{counterfactual explanation}) with the question "\textit{what is the slightest change I can make on the input graph to achieve a desired prediction?}". 

\textbf{Factual explanation.}
With its ultimate goal, factual explanation naturally relates to generalization w.r.t. the common focus on the rationale that causes the prediction target. This concept is intuitively equivalent to the invariant variables in invariant learning, or causal variables in causal models that are stable across environments. Therefore, many explanation methods also naturally have a dual objective in generalization, including invariant learning methods such as DIR \cite{DIR}, GSAT \cite{GSAT}, \textbf{GREA} \cite{GREA_2022}; causal modeling methods like CAL/CAL+ \cite{CAL_2022,CAL+_2024} and DSE \cite{DSE_2022}, and Granger causality \cite{granger1969investigating} based methods such as \textbf{GEM} \cite{GEM_2021} and \textbf{CI-GNN} \cite{CI-GNN_2024}.



\textbf{Counterfactual explanation.} 
Counterfactual explanation \cite{wachter2017counterfactual} studies the problem of making perturbations on the input graph to change the model prediction. 
Even though the original concept of counterfactual explanation does not involve causal inference, recent studies \cite{mahajan2019preserving,ma2022clear} have started to discuss the benefit of explicitly incorporating causality into counterfactual explanation. In this context, it is worth noting that making perturbations itself would result in certain outcomes, which relates to counterfactual reasoning in a causal context. Therefore, many recent works utilize causal methods for graph counterfactual explanation such as \textbf{CLEAR} \cite{ma2022clear}, which involves a deep graph generative model to promote causality in counterfactual explanation. Another work \textbf{CF$^2$} \cite{CF2_2022} combines both factual reasoning and counterfactual reasoning to obtain the necessary explanation for graph models.

\subsection{Robustness}
Adversarial robustness refers to the ability of a model to perform correctly and maintain its integrity under adversarial attacks. These attacks are typically slight perturbations on the original input data, also known as adversarial examples. These adversarial examples are usually imperceptible by human eyes but are crafted to mislead the model into making mistakes. Adversarial robustness and OOD generalization are interconnected areas as the adversarial samples can be considered from an adversarial distribution outside of the training data. Explorations in this area from a causal view are rare, but still offer valuable insights. For example, \textbf{IDEA} \cite{IDEA_2024} considers the causal features as the key to graph adversarial robustness due to their invariance across attacks. Based on this, it designs both node-based and structure-based invariance objectives to capture causally invariant node representations against adversarial attacks. Under linear causal assumptions, the defense strategy of IDEA has been proved to be causally invariant.

\subsection{Fairness}
Fairness in machine learning aims to eliminate the bias against any demographic groups or individuals. It is widely regarding certain sensitive features (e.g., age, gender). There have been lots of notions of fairness defined from different perspectives. In recent years, causality-based fairness has attracted more and more attention since it can track the root and path of bias in its generation process, providing explanatory and controllable approaches for mitigating potential discrimination. Counterfactual fairness \cite{kusner2017counterfactual} is one of the most well-known notions in this line, which measures fairness by comparing the model prediction under one's original sensitive feature and a counterfactual case. An example question is: would a male applicant have the same chance to get a job offer if he had been a female?

A fair model must effectively handle various sensitive feature values, which are often regarded as different domains. In this sense, a fair model should also demonstrate strong generalization capabilities across these different domains. From a causal view, it requires identifying the causal path from sensitive features to other variables used in prediction, and eliminating (a part of) these paths to achieve fairness. This relates to the mitigation of environment effects on model prediction in generalization. On graphs, fairness is a more complicated task due to the bias (causally) propagated through graph links. Studies for counterfactual fairness, such as \textbf{GEAR} \cite{ma2022clear} and \textbf{RFCGNN+} \cite{RFCGNN+_2024} incorporate causal reasoning in graphs.

\section{Future Work}
\label{sec:future}
In this paper, we have conducted a comprehensive review of causality-involved approaches for OOD generalization of graph machine learning. We begin by highlighting the motivations and challenges inherent to this area, providing a structured categorization of existing methodologies based on their technical approaches. Furthermore, we discuss the commonalities and differences of methods in this field, and also introduce the connections between them and related studies in other areas. Through this analysis, we have extracted valuable insights and established a robust foundation for ongoing exploration in related fields.

Looking ahead, there are several promising avenues for further studies:

\begin{itemize}
    \item \textbf{Application in high-stakes domains for graph trustworthiness}: Causal models in important domains such as science, finance, law, and health often incorporate domain knowledge and require a high degree of trustworthiness. These fields demand rigorous frameworks when modeling the causal relationships as well as the graph structure due to their reliance on precise and professional knowledge. Future research could focus on customizing domain-specific causality-enhanced graph models to maintain the trustworthiness and application of GML in high-stakes domains.
    \item \textbf{Uncertainty quantification of causality-involved GML}: While existing GML methods have demonstrated impressive performance, it is crucial to include uncertainty quantification, especially under distribution shifts on graphs. This aspect might intersect with emerging research on conformal prediction in causal inference \cite{lei2021conformal} and graph learning \cite{zargarbashi2023conformal}. Addressing uncertainty quantification will enhance the reliability and applicability of GML in varying environments.
    \item \textbf{Graph-related AGI incorporating causal knowledge}: Another exciting direction is the integration of causal knowledge into graph foundation models (GFMs). GFMs have the capability of addressing a broader range of tasks beyond traditional GML limits. While large models exhibit superior generalization, they still face challenges such as bias from training domains. Incorporating human-provided or data-driven causal insights into these foundation models has the potential to effectively improve their trustworthiness and mimic complex human reasoning processes, which represents a significant step toward achieving artificial general intelligence (AGI). But currently, this task is much more challenging than traditional causality-involved  GML due to the large scale of GFMs. 
\end{itemize}

\section{Acknowledgements}
The authors have no conflicts of interest to report.